\documentclass[10pt,twocolumn,letterpaper]{article}

\usepackage[pagenumbers]{cvpr} 

\definecolor{cvprblue}{rgb}{0.21,0.49,0.74}
\usepackage[pagebackref,breaklinks,colorlinks,allcolors=cvprblue]{hyperref}
\definecolor{ratioae}{rgb}{0.91,0.90,0.98}
\definecolor{ratioaeDeep}{HTML}{483D8B}  

\def\confName{CVPR}
\def\confYear{2027}

\usepackage{amsmath}
\usepackage{amsthm}
\usepackage{bm}
\usepackage{threeparttable}
\usepackage{multirow}
\usepackage{makecell}
\usepackage{subcaption}
\usepackage{transparent}
\usepackage[dvipsnames]{xcolor}
\usepackage[table]{xcolor}
\allowdisplaybreaks

\theoremstyle{plain}
\newtheorem{proposition}{Proposition}[section]

\title{AdaS-ViT: Test-Time Adaptive Routing for Spiking Vision Transformers with Ultra-High Energy Efficiency}

\author{
    Zecheng Hao\textsuperscript{1,2,4}\hspace{0.15em} Zihan Huang\textsuperscript{1,2,4}\hspace{0.15em} Shenghao Xie\textsuperscript{1,2}\hspace{0.15em} Jiyuan Zhang\textsuperscript{3}\thanks{Corresponding author.}\hspace{0.4em} Zhaofei Yu\textsuperscript{1,4}\hspace{0.15em} Tiejun Huang\textsuperscript{1,2,3} \\
    \textsuperscript{1}School of Computer Science, Peking University\\
    \textsuperscript{2}State Key Laboratory for Multimedia Information Processing, Peking University\\
    \textsuperscript{3}Beijing Academy of Artificial Intelligence (BAAI)\\
    \textsuperscript{4}Beijing Key Laboratory of Brain-inspired Spiking Large Models, Peking University\\
    {\fontsize{8.5}{\baselineskip}\selectfont\ttfamily\{haozecheng,yuzf12,tjhuang\}@pku.edu.cn,\{hzh,shenghaoxie\}@stu.pku.edu.cn,jyzhang01@baai.ac.cn}
}

\begin{document}
\maketitle
\begin{abstract}
Spiking Neural Networks (SNNs) promise event-driven and energy-efficient vision, yet existing spiking vision transformers (S-ViTs) largely confine spiking tokens to fixed and uniform firing schemes along the temporal, spatial and channel dimensions, whereas suppressing the computational redundancy thus incurred calls for adaptive firing mechanisms that operate jointly across all three axes at test time. To this end, we propose the Adaptive Efficient Integrate-and-Fire (AdaE-IF) model, which consults the attention score of each spiking token and routes salient regions to fine-grained firing patterns while assigning energy-frugal coarse patterns to the remaining regions. Building on AdaE-IF, we further introduce Spiking Gated Self-Attention (S-GSA) and Spiking Local-Gated Unit (S-LGU), which enforce selective spike emission at token and channel granularity. We integrate the aforementioned mechanisms and modules into AdaS-ViT, a test-time adaptive-routing architecture trained with a progressive spatio-temporal curriculum that stabilizes the co-adaptation of firing precision and routing masks. Extensive experiments across benchmarks spanning diverse data scales and modalities demonstrate that AdaS-ViT consistently achieves highly competitive performance. Notably, on ImageNet-1k it consumes only 1.02 to 2.45 millijoules per image and attains an accuracy-per-energy ratio of 32× to 72×, operating in an economical inference regime and thereby pushing the energy-efficiency boundary further.
\end{abstract}    
\section{Introduction}
\begin{figure} [t]\centering  
\includegraphics[width=\columnwidth]{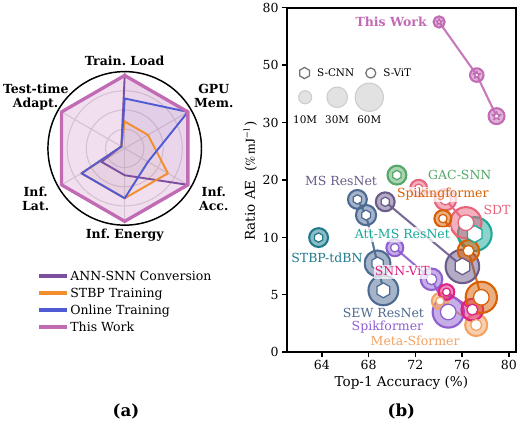}
\caption{
  \textbf{AdaS-ViT establishes a compelling accuracy-energy frontier for spiking vision transformers.}
  \textbf{(a)} Qualitative comparison of training paradigms over six axes covering training cost and inference behavior: prior paradigms trade these objectives against one another, whereas our test-time adaptive routing maintains a uniformly strong profile across all of them.
  \textbf{(b)} Quantitative comparison with prior SNN-based works on ImageNet-1k. The horizontal axis denotes top-1 accuracy and the vertical axis reports Ratio AE, i.e., the accuracy delivered per millijoule of inference energy. AdaS-ViT occupies the top-right region alone with the smallest bubble area, indicating markedly higher energy efficiency at a fraction of the parameters of prior art.
  }
  \label{fig01}      
\end{figure}
As a brain-inspired neural network paradigm \cite{maas1997networks}, Spiking Neural Networks (SNNs) emit spikes selectively according to the input current, memorized membrane potential and firing threshold, therefore exhibit remarkable energy-efficient advantage when deployed on neuromorphic hardware \cite{merolla2014million, ma2017darwin, davies2018loihi, debole2019truenorth, pei2019towards}, holding tremendous potential for mobile computing and low-power inference scenarios. However, pushing such an advanced architecture, in particular the spiking vision transformer (S-ViT), toward ultra-high inference energy efficiency still faces a series of fundamental academic challenges and performance bottlenecks.

Current mainstream learning frameworks for SNNs mainly include ANN-SNN Conversion \cite{cao2015spiking, bu2022optimal, huang2024towards, hao2025conversion}, Spatio-Temporal Back-propagation (STBP) \cite{wu2018STBP} and Online Training \cite{Meng2023SLTT}, as illustrated in Fig.\ref{fig01}(a). Among them, the scheme of conversion learning first trains a native or quantization-aware Artificial Neural Network (ANN) and then maps it into an equivalent SNN at inference time, exploiting the latent mathematical equivalence between ANN activations and average spike firing rates. In theory, this paradigm inherits a constant training cost comparable to that of ANNs together with a high accuracy ceiling. In practice, however, since it only considers the globally averaged firing behavior over the temporal dimension while ignoring the temporal information carried by spike sequences, conversion errors accumulate progressively across layers during inference stage \cite{hao2023reducing}. Consequently, far more time-steps than those used in training-time simulation are required to recover the accuracy of the converted SNN, incurring substantial inference latency and energy consumption.

STBP Training, in contrast, unfolds gradient propagation over both the spatial and temporal dimensions in a manner analogous to Recurrent Neural Networks (RNNs). While this framework eliminates the performance gap between training and inference, it must resort to surrogate gradients for the non-differentiable spike firing process \cite{neftci2019surrogate}. The resulting gradient inconsistency is progressively amplified along the spatio-temporal dimensions, capping the learning capability of SNNs. As a result, SNNs fail to exhibit a clear scaling law when the number of training time-steps is increased, while GPU memory grows linearly with the simulated steps and may even explode. To remedy these deficiencies of vanilla STBP, Online Training has been proposed as a compromise: by detaching gradients along the temporal dimension, it retains the standard forward process of spiking neurons while enjoying ANN-like constant training memory, constituting a lightweight scheme that supports low-latency inference. Nevertheless, on top of the gradient inconsistency inherent to spike emission in STBP, it further truncates the gradients of the membrane-potential accumulation, thereby compressing the accuracy ceiling of SNNs even further.

Compared with single-path and sequentially-computed architectures such as spiking convolutional neural networks (S-CNNs), S-ViTs exhibit far more intricate spike-sequence flow paths and additionally require spiking neuron models to emulate floating-point multiplicative modules (\eg, Softmax), which poses severe challenges to the inference accuracy of the aforementioned traditional learning paradigms under low-latency constraints. Moreover, none of these methods can effectively regulate the number of emitted spikes, seriously hindering the progress of S-ViTs toward the core research goal of ultra-low inference power. Recently, bit-level conversion training for spike sequences \cite{shen2025spikepack} has offered a preliminary attempt to break this dilemma: by establishing a mapping between spike sequences and quantized activations from the perspective of numerical radix representation, one can decide, according to the number of spikes associated with each quantization level, whether to assign that level to an energy-efficient activation subset \cite{li2020additive, hao2026svit}.

Notably, all of the above schemes share one critical deficiency: they apply an entirely identical and rigid computational mechanism to every spiking feature map. On the one hand, this prevents the learning framework from performing truly precision-level regulation at the granularity of individual spiking tokens or features; on the other hand, spiking features of unequal importance are allocated firing-pattern sets of identical capacity, causing latent redundant computation across multiple dimensions and obstructing further improvements in the energy-efficiency ratio of S-ViTs. In this work, we pioneer the notion of precision-level adaptive computation for S-ViTs, achieving redundancy-free, energy-efficient inference across all dimensions at test time. Our contributions are summarized as follows:
\begin{itemize}
    \item We propose Adaptive Efficient Integrate-and-Fire (AdaE-IF) model, an advanced spiking neuron that adaptively routes each spiking token to a distinct spike firing pattern according to its importance, simultaneously possessing ultra-high energy efficiency and fine-grained representation capability. On this basis, we further establish a unified energy-consumption calculation standard for spiking models and quantitatively demonstrate the mechanistic advantages of AdaE-IF.
    \item We design the novel Spiking Gated Self-Attention (S-GSA) and Spiking Local-Gated Unit (S-LGU), which not only enable element-level computational regulation along the token and feature dimensions but also effectively suppress redundant computation over spike sequences.
    \item Integrating the above modules and mechanisms, we propose the AdaS-ViT architecture, which innovatively endows S-ViTs with test-time adaptive inference and computation along the temporal, spatial and channel dimensions simultaneously. A dual progressive scheduling strategy over token-routing and activation-quantization ratios guarantees stable curriculum learning of AdaS-ViT.
    \item Across multiple key evaluation metrics, including parameter count, energy, accuracy and Ratio AE, AdaS-ViT exhibits superior performance on a series of challenging static and neuromorphic benchmarks.
\end{itemize}
\section{Related Works}
\textbf{Learning paradigms of SNNs.} ANN-SNN Conversion \cite{cao2015spiking} can be categorized into several sub-families, including training-free conversion \cite{li2021free}, conversion-simulated training \cite{bu2022optimal}, post-hoc correction of conversion errors \cite{hao2023reducing} and parallel conversion \cite{hao2025conversion}, which aim to directly obtain a converted model equipped with SNN-style inference capability either at a constant training cost or by re-deploying neuron layers upon customized ANNs. In the conventional circumstance that employs vanilla spiking models, conversion-based learning strategies usually fail to rapidly recover inference performance within a small number of time-steps. In comparison, surrogate-gradient-based STBP Training \cite{wu2018STBP, neftci2019surrogate} has become an effective remedy for this issue. By incorporating related modules \cite{zheng2021going}, training strategies \cite{deng2022temporal} and spiking computation mechanisms \cite{huang2024clif} that enhance the exploitation of temporal information, researchers have further optimized the representation capability of STBP-based SNNs under low-latency conditions. To address the problem that the training memory of STBP-type methods grows linearly with the number of time-steps, Online Training \cite{Meng2023SLTT, jiang2024ndot} explores a trade-off route that precisely simulates the spiking forward computation while approximating the back-propagated spatio-temporal surrogate gradients. In addition, ANN-SNN knowledge distillation \cite{xu2023constructing}, SNN self-distillation \cite{ding2025rethinking} and spike-timing-based training algorithms \cite{zhu2023exploring} have also offered the SNN community additional academic inspiration from diverse perspectives.

\textbf{STBP-based S-ViTs.} As a pioneering effort in the S-ViT domain, Spikformer \cite{zhou2023spikformer} first built a fully spiking self-attention (SSA) computation module. Building upon this, SDT \cite{yao2023attention} with linear computational complexity, Meta-Sformer \cite{yao2024spike} and S-Resformer \cite{shi2024spikingresformer} that hybridize convolution and attention mechanism, Spikingformer \cite{zhou2026spikingformer} with membrane-based shortcuts have been proposed in succession. Moreover, the similarity-measurement issue in vanilla SSA computation has been further refined \cite{wang2025svit, xiao2025rethinking}. Meanwhile, researchers have pointed out that skillfully integrating spiking models with diverse firing mechanisms into SSA can strengthen its representation capability while preserving the addition-only computation property \cite{guo2025spiking}. QKformer \cite{zhou2024qkformer} explores a lightweight attention computation mechanism based on a dual-branch structure, whereas Maxformer \cite{fang2025maxformer} highlights the important value of high-frequency information in the learning process of S-ViTs.

\textbf{Efficient learning for S-ViTs.} Obtaining high-performance S-ViTs within a constant training overhead has long been a central concern in the SNN learning community. Combining conventional conversion learning with S-ViTs is a straightforward idea \cite{wang2023masked, huang2024towards}, yet it is constrained by the inherent structure and computation pipeline of the pre-trained ViT itself: for instance, modules involving non-SNN operators usually require tedious fitting with high-precision spiking models, while conversion schemes based on vanilla spiking models need a large number of time-steps to approximately restore the inference accuracy to that of the corresponding reference ANN-based ViTs \cite{dosovitskiy2021vit, liu2021swin}. To break through the above limitations, researchers have respectively proposed equivalent conversion training schemes based on integer spikes \cite{yao2025scaling, xie2026SAFformer} and spike firing patterns \cite{shen2025spikepack, hao2026svit}, which effectively strengthen the representation fidelity of the forward process and the gradient consistency of the backward process while guaranteeing leading low-latency inference performance.
\section{Preliminaries}
\textbf{LIF \& IF model} \cite{gerstner2002spiking} can be regarded as a Markov model whose basic state is $(\lambda^l,\mathbf{v}^l,\mathbf{I}^l,\mathbf{s}^l,\theta^l_\text{Pre},\theta^l_\text{Post})$ and whose state-transition process consists of charging, firing and resetting. Here $\lambda^l$ denotes the membrane leakage factor, $\mathbf{W}^l$ is the synaptic-layer weight, and $\mathbf{I}_t^l=\mathbf{W}^l\mathbf{s}_t^{l-1}\theta_\text{Post}^{l-1}$ represents the input current. Once the accumulated membrane potential $\mathbf{m}_t^l=\lambda^l\mathbf{v}_{t-1}^l+\mathbf{I}_t^l$ at the current time-step reaches the firing threshold $\theta_\text{Pre}^l$, the model emits spikes toward the post-synaptic layer, \ie, $\mathbf{s}_t^l:=(\mathbf{m}_t^l\geq \theta_\text{Pre}^l)$, where $\theta^l_\text{Post}$ serves as a scaling factor applied to the spike firing sequence $\{\mathbf{s}_t^l \}$. The accumulated membrane potential is then restored, through the resetting process, to the post-firing membrane potential $\mathbf{v}_t^l=\mathbf{m}_t^l-\mathbf{s}_t^l\theta_\text{Post}^l$. The overall computation can be consolidated into Eq.\eqref{eq01}.
\begin{align}
    \mathbf{v}_t^l &= \lambda^l\mathbf{v}_{t-1}^l+\mathbf{I}_t^l - \mathbf{s}_t^l\theta_\text{Post}^l, \label{eq01} \\
    \tilde{\mathbf{F}}_{1:T}^l &= \frac{1}{T}\sum_t \mathbf{s}_t^l\theta_\text{Post}^l
    = \frac{1}{T}\sum_t\mathbf{I}_t^l + (\lambda^l\mathbf{v}_{t-1}^l - \mathbf{v}_t^l) \nonumber \\
    &\overset{\ast}{=} \tilde{\mathbf{I}}_{1:T}^l + \frac{\mathbf{v}_0^l - \mathbf{v}_T^l}{T}. \label{eq02}
\end{align}
\textbf{Firing rate-Input current function} offers an insightful perspective for studying SNN learning algorithms. By accumulating Eq.\eqref{eq01} along the temporal dimension, we obtain the expression that relates the average firing rate $\tilde{\mathbf{F}}_{1:T}^l$ to the average input current $\tilde{\mathbf{I}}_{1:T}^l=\mathbf{W}^l\tilde{\mathbf{F}}_{1:T}^{l-1}$, where the condition $\ast$ refers to $\lambda^l=1$. In other words, for the IF model, the spike firing rates of adjacent layers approximately obey a linear transformation, which constitutes the theoretical foundation for the feasibility of ANN-SNN Conversion. In particular, when $\{\mathbf{I}_t^l\}$ follows a perfectly uniform distribution, the only remaining temporal-information error term $(\mathbf{v}_0^l-\mathbf{v}_T^l)/T$ in Eq.\eqref{eq02} also becomes a deterministic value, rendering the transformation from a customized ANN to the converted SNN entirely lossless. In this case, $\tilde{\mathbf{F}}_{1:T}^l$ and $\tilde{\mathbf{I}}_{1:T}^l$ follow the staircase-shaped curve shown in Fig.\ref{fig02}. Conversely, for the LIF model ($\lambda^l\in(0,1)$) or when the input current is non-uniformly distributed, the above direct-mapping-style training scheme becomes less feasible, and one must then resort to STBP Training.

\textbf{Energy-budget calculation.} Building on the above discussion, Eq.\eqref{eq03} establishes a general formulation for the energy consumption of spiking models from the perspective of mathematical expectation. When $\tilde{\mathbf{F}}_{1:T}^l$ and $\tilde{\mathbf{I}}_{1:T}^l$ satisfy the ideal staircase-shaped relationship, Eq.\eqref{eq03} can be further transformed into a form that supports piecewise integration. Here $e(\cdot)$ is the calculation function for energy consumption, $p(\cdot)$ is the probability distribution function over the input current, while $[\mathbf{I}_{1:T}^{l,(i)}, \mathbf{I}_{1:T}^{l,(i+1)})$ represents the $i$-th segment interval of $\tilde{\mathbf{I}}_{1:T}^l$.
\begin{align}
    \mathcal{E}_{1:T}^l &= \int \!e\! \left(\tilde{\mathbf{F}}_{1:T}^l|\{\mathbf{I}_t^l\} \right) p\!\left(\{\mathbf{I}_t^l\}(x) \right) dx \nonumber \\
    &= \sum_i \int_{\mathbf{I}_{1:T}^{l,(i)}}^{\mathbf{I}_{1:T}^{l,(i+1)}} \!\! e\!\left(\tilde{\mathbf{F}}_{1:T}^l(\tilde{\mathbf{I}}_{1:T}^l) \right) p\!\left(\tilde{\mathbf{I}}_{1:T}^l(x) \right) dx.
    \label{eq03}
\end{align}
\section{Methods}
\subsection{AdaE-IF}
\begin{figure}[t]
  \centering
  \includegraphics[width=\columnwidth]{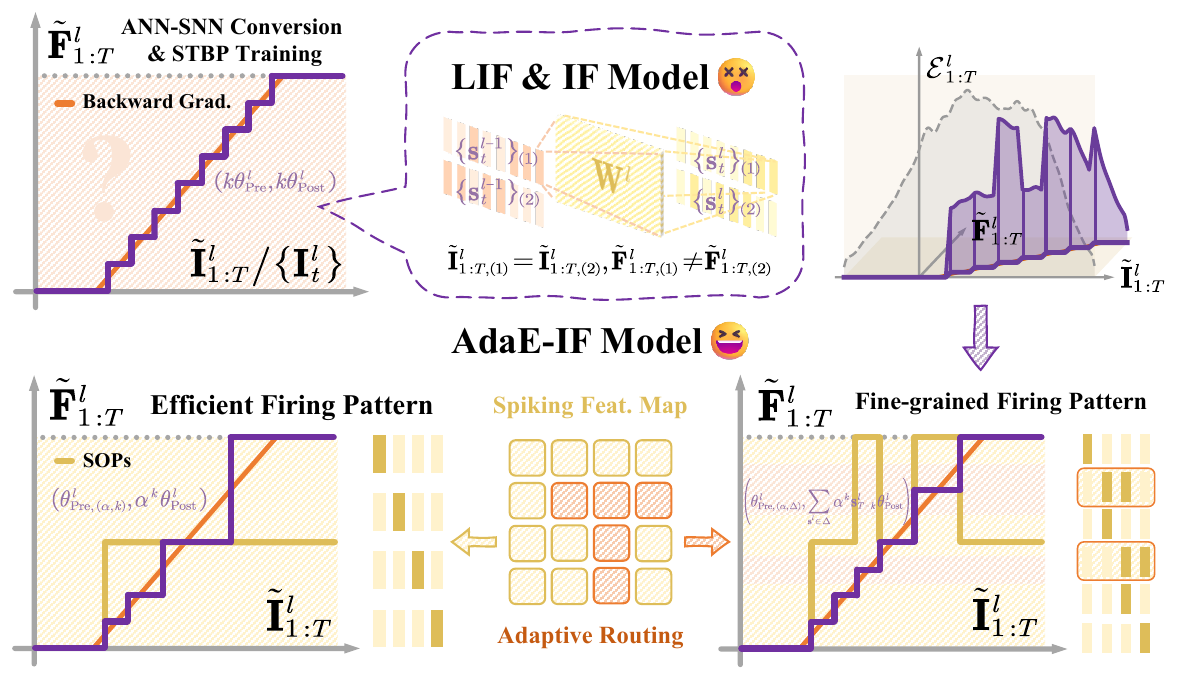}
  \caption{\textbf{AdaE-IF: from uniform to adaptive and efficient spike firing.}
  Top-left: the firing-rate curve of a vanilla IF/LIF neuron is ideally a uniform staircase, yet diverse input current patterns $\{ \mathbf{I}_t^l \}$ yield divergent firing rates $\tilde{\mathbf{F}}_{1:T}^l$ for identical average input current $\tilde{\mathbf{I}}_{1:T}^l$, which accumulate across layers into conversion error and spatio-temporal gradient inconsistency.
  Bottom: AdaE-IF maps each spike firing pattern to its bit-granular equivalent, aligning training gradients and enabling precise regulation of Synaptic Operations (SOPs); guided by the attention score of the spiking feature map (Middle), it routes each token to a coarse-grained pattern that preserves energy efficiency, or a fine-grained pattern that enhances representation at only marginal extra cost; jointly scheduled under an ultra-low energy budget (Top-right), the two modes realize test-time adaptive firing precision.}
  \label{fig02}
\end{figure}
\begin{figure*}[t]
  \centering
  \includegraphics[width=\textwidth]{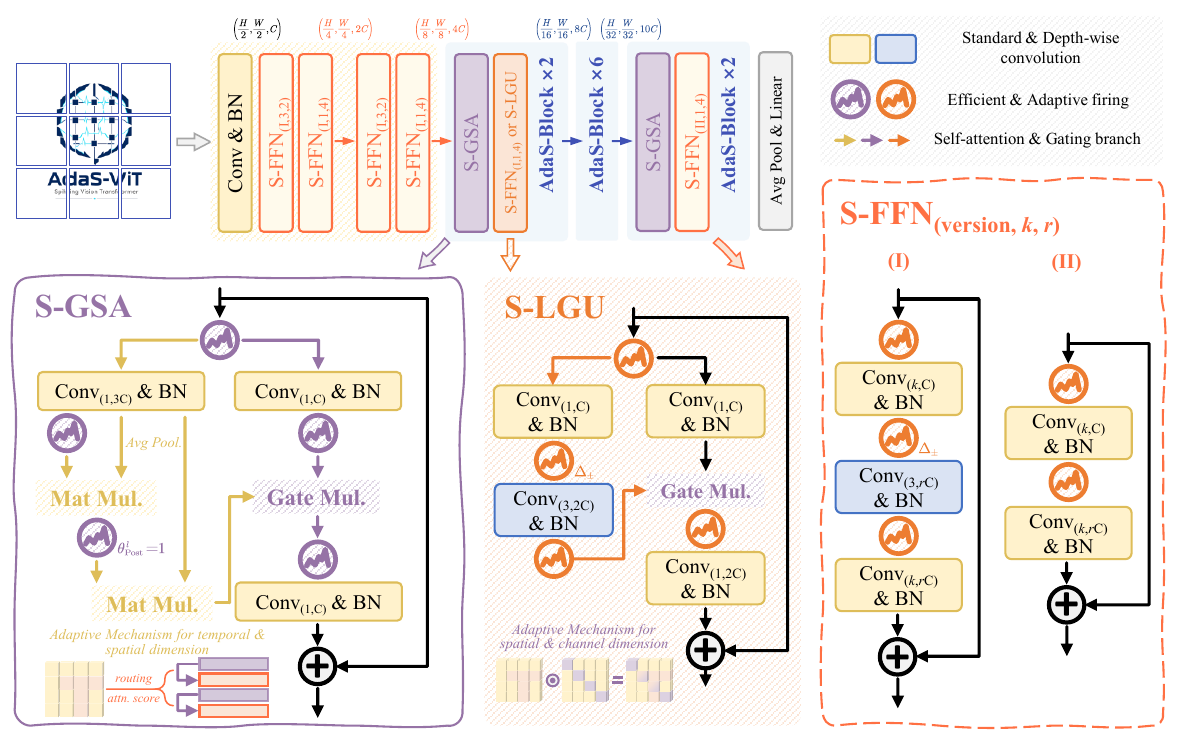}
  \caption{\textbf{Architecture of AdaS-ViT.}
  Top: the macro pipeline stacks dual-mode AdaS-Blocks (S-GSA followed by S-LGU or S-FFN) over progressively downsampled token maps.
  Bottom: data flow of S-GSA, S-LGU and the parameterized S-FFN variants $(\mathrm{I})(\mathrm{II})$ with convolution kernel size $k$ and expansion ratio $r$.}
  \label{fig03}
\end{figure*}
As discussed above, existing ANN-SNN Conversion and STBP Training schemes built upon the vanilla IF \& LIF model do not satisfy the condition that $\tilde{\mathbf{F}}_{1:T}^l$ and $\tilde{\mathbf{I}}_{1:T}^l$ exhibit a staircase-shaped relationship, which prevents them from performing exact gradient back-propagation in the form of $\text{ClipReLU}(\cdot)$ and thus from achieving effective SNN learning in an ANN-like manner. Moreover, as the value of $\tilde{\mathbf{F}}_{1:T}^l$ increases, the corresponding number of emitted spikes under such a conventional mapping also grows, which severely hinders the joint optimization of precise representation and high energy efficiency in SNNs. To address the above issues, a bit-level mapping scheme can be considered for spike sequences: $\mathbf{s}^l=\{\alpha^{T-1}\mathbf{s}_1^l,\cdots,\alpha\mathbf{s}_{T-1}^l, \mathbf{s}_T^l \}, \mathbf{s}_t^l\in\{0,1,\cdots,\alpha\!-\!1 \}$. Here $\alpha=2$ is typically adopted to remain consistent with the low-level numerical computation mechanism of the hardware, so that during SNN inference the spikes at different firing time-steps can be converted into one another with merely a single shift operation.

Intuitively, under the bit-level direct mapping, the number of emitted spikes is already reduced from $O(\tilde{\mathbf{F}}_{1:T}^l)$ to $O(\log_{\alpha} \tilde{\mathbf{F}}_{1:T}^l)$, and the linear correspondence between the two quantities is thereby decoupled. Building upon this, a more energy-efficient scheme can be further considered: if a subset $\Delta$ of spike firing patterns is manually selected, \eg, $\Delta=\{\mathbf{s}_1^l,\cdots,\mathbf{s}_{T-1}^l, \mathbf{s}_T^l \}, \alpha=2, \sum\mathbf{s}_t^l\leq n$, then under this subset constraint the number of emitted spikes corresponding to every admissible firing pattern is upper-bounded by $n$.

Nevertheless, it should be noted that a trade-off exists between representation capability and energy efficiency, since a smaller $n$ is obviously not always preferable. When $n=1$, from the perspective of activation quantization, the mapping between $\tilde{\mathbf{F}}_{1:T}^l$ and $\tilde{\mathbf{I}}_{1:T}^l$ directly degenerates from fully uniform quantization into exponential quantization, making the training process highly prone to collapse. Therefore, as shown in Fig.\ref{fig02}, we propose the AdaE-IF model, an extreme-point solution that pursues both high representation capability and energy efficiency: by selectively triggering spike firing patterns of different granularities for spiking tokens of different importance, it maximizes computational power savings while stabilizing the learning capability of the model. Specifically, we sequentially pick the corresponding firing patterns in ascending order of $\sum\mathbf{s}_t^l$. Under this selection strategy, the efficient firing pattern terminates earlier whereas the fine-grained firing pattern terminates later, so that the efficient firing pattern is inherently more energy-frugal while the fine-grained firing pattern is more precise. By comparing the various spike firing mechanisms discussed above, we arrive at the following conclusion:
\begin{proposition}[Upper-bound of energy-budget under different spike firing mechanisms]
\label{prop:budget}
According to the piecewise-integration calculation standard in Eq.\eqref{eq03}, we can have 
$\mathcal{E}_{1:T}^{l,\text{IF}} \geq \mathcal{E}_{1:T}^{l,\text{Bit}} \geq \mathcal{E}_{1:T}^{l,\text{Bit-Pick}} \geq \mathcal{E}_{1:T}^{l,\text{AdaE-IF}}$, here IF, Bit and Bit-Pick respectively refer to spiking models under the condition of vanilla conversion, bit-level mapping and selecting the subset of firing pattern.
\end{proposition}
To strengthen the stability of the AdaE-IF model during training, we devise a progressive learning scheme for $\tilde{\mathbf{F}}_{1:T}^l$ along two dimensions, namely the firing-pattern quantization ratio $r_\text{fp}$ and the token-routing ratio $r_\text{tok}$:
\begin{align}
    \tilde{\mathbf{F}}_{1:T}^l \!=\! (1 \!-\! r_\text{fp}) \tilde{\mathbf{F}}^l_{\!(0,\theta_\text{Post}^l)\!} \!+\! r_\text{fp}\! \left(\! (\! \mathbf{1} \!-\! \mathbf{m}_\text{tok}^l \! ) \tilde{\mathbf{F}}_{1:T}^{l,\text{eff}} \!+\! \mathbf{m}_\text{tok}^l  \tilde{\mathbf{F}}_{1:T}^{l,\text{fine}} \!\right)\!.
\end{align}
Here $\mathbf{m}_\text{tok}$ denotes the routing mask over spiking tokens, whose masking ratio equals $r_\text{tok}$. $\tilde{\mathbf{F}}_{(0,\theta_\text{Post}^l)}^l=\text{clamp}(\cdot,0,\theta_\text{Post}^l)$ and $r_\text{fp},r_\text{tok}$ are scheduled according to a cosine annealing strategy as the number of training epochs increases. $\tilde{\mathbf{F}}_{1:T}^{l,\text{eff}}$ and $\tilde{\mathbf{F}}_{1:T}^{l,\text{fine}}$ correspond to the total firing amounts of the weighted spike sequences under the two aforementioned firing patterns, respectively. In other words, in the numerical dimension AdaE-IF manifests as a hybrid smoothing of $\text{ClipReLU}(\cdot)$ and the spike firing pattern, while within the firing pattern itself it is initialized with efficient firing and progressively introduces fine-grained firing for salient regions as the learning capability of the SNN evolves.

\subsection{Spiking Gated Computation Modules}
The fusion between gating mechanisms and attention or feed-forward computation has been recognized by prior studies in the ANN community as holding great potential \cite{liu2021gMLP, qiu2025gated}, but how to effectively integrate such gating into SNNs while supporting spiking-style computation remains an open problem. In the attention module, since SNNs cannot support Softmax-related operators and typically substitute them with spiking neuron layers, the divergent attention-score distribution and the redundant computational overhead in vanilla SSA deserve further research. To this end, as illustrated in Fig.\ref{fig03}, we propose S-GSA, which introduces a dual-branch spiking gating computation mechanism to effectively regulate the attention paid to salient regions and, together with the selective spike firing mechanism, reinforces the energy-efficient nature of SSA computation. For the Spiking Feed-forward Network (S-FFN), we combine the temporal-dimension tunability of AdaE-IF with the spatial- and channel-dimension controllability of the spiking gating module, then further derive S-LGU, which possesses multi-dimensional and energy-efficient adaptive computation capability.

\textbf{S-GSA.} Given the input spike sequence of the module $\mathbf{X}_\text{in}^l\in\mathbb{R}^{TB\times C\times H\times W}$, we perform standard self-attention computation based on the efficient firing pattern $\text{Eff-IF}(\cdot)$ and obtain $\mathbf{Attn}^l\in\mathbb{R}^{TB\times h\times HW\times HW}$, where $B,h,H,W$ denote the batch size, the number of attention heads, the height and width of the spiking feature map, respectively. Meanwhile, $\mathbf{X}_\text{in}^l$ also passes through the gating branch and is combined with the spike flow after attention projection via a Hadamard product to yield the sparsified fused representation $\mathbf{X}_\text{gate}^l\in\mathbb{R}^{TB\times C\times H\times W}$, which then flows through the synaptic layer and the residual connection to produce the output current $\mathbf{Y}^l\in\mathbb{R}^{TB\times C\times H\times W}$.
\begin{align}
    \mathbf{X}_\text{in}^l &= \text{Eff-IF}(\mathbf{X}^l), \\
    \mathbf{Attn}^l &= \text{Eff-IF}\left( \text{Eff-IF}(\mathbf{W}_\text{Q}^l\mathbf{X}_\text{in}^l) \mathbf{W}_\text{K}^l\mathbf{X}_\text{in}^l \right), \\
    \mathbf{X}_\text{gate}^l &= \mathbf{Attn}^l(\mathbf{W}_\text{V}^l\mathbf{X}_\text{in}^l) \odot \text{Eff-IF}(\mathbf{W}_\text{G}^l\mathbf{X}_\text{in}^l), \\
    \mathbf{X}_\text{out}^l &= \mathbf{W}_\text{O}^l \text{Eff-IF}(\mathbf{X}_\text{gate}^l), \\
    \mathbf{Y}^l &= \mathbf{X}_\text{out}^l + \mathbf{X}^l.
\end{align}
\textbf{S-LGU.} Here $\mathbf{X}_\text{in}^l$ flows into the standard linear transformation branch and the local feature computation branch, respectively. Specifically, the local feature extraction sequentially invokes the point-wise and depth-wise convolutions $\mathbf{W}_{\text{G}_1}^l,\mathbf{W}_{\text{G}_2}^l$, then performs the gated activation via AdaE-IF. $\Delta_{\pm}$ denotes that the AdaE-IF employed here adopts a spike firing pattern symmetric in positive and negative values.
\begin{align}
    \mathbf{X}_\text{in}^l &= \text{AdaE-IF}(\mathbf{X}^l), \\
    \mathbf{X}_\text{feat}^l &= \text{AdaE-IF} \left(\mathbf{W}_{\text{G}_2}^l \text{AdaE-IF}(\mathbf{W}_{\text{G}_1}^l \mathbf{X}_\text{in}^l)_{\!\Delta_{\pm}\!} \right), \\
    \mathbf{X}_\text{gate}^l &= \mathbf{W}_\text{I}^l\mathbf{X}_\text{in}^l \odot \mathbf{X}_\text{feat}^l, \\
    \mathbf{X}_\text{out}^l &= \mathbf{W}_\text{O}^l \text{AdaE-IF}(\mathbf{X}_\text{gate}^l), \\
    \mathbf{Y}^l &= \mathbf{X}_\text{out}^l + \mathbf{X}^l.
\end{align}

\subsection{AdaS-ViT}
\textbf{AdaS-Block.} We further integrate the above modules into a unified Transformer block. For each AdaS-Block, we assign one S-GSA and select one S-LGU or S-FFN, where the specific selection and configuration strategy of the modules depends on the stage in which the corresponding block resides. As shown in Fig.\ref{fig03}, the overall pipeline of AdaS-ViT is partitioned into multiple stages and the number of spiking tokens is progressively compressed as the stages evolve. In particular, Stage I performs preliminary information extraction and compression solely by stacking multiple convolution-extended S-FFN variants. For Stage II \& III, we perform pre-pooling on $\mathbf{W}_\text{K}^l,\mathbf{W}_\text{V}^l$ in S-GSA to further save computational overhead, as well as adopting an interleaved usage scheme for S-LGU and $\text{S-FFN}_\text{(I)}$ so as to comprehensively memorize the global and local knowledge representations learned by AdaS-ViT. As for Stage IV, since the spiking feature map at this point has been down-sampled to an extremely small size, no further compression or local information extraction is needed. We remove the pooling operation of S-GSA and the introduction of the S-LGU module, then uniformly employ $\text{S-FFN}_\text{(II)}$.
\begin{align}
    &\text{AdaS-B}(\cdot) = \left\{ \text{S-GSA}^l, \{ \text{S-LGU}, \text{S-FFN} \}^{l+1} \right\}, \\
    &\text{AdaE-IF}.\mathbf{m}_\text{tok}^{l+1} = \mathbf{1}\! \left[\arg\!\!\!\max_{r_\text{tok}\times HW}\!\!\! \tilde{\mathbf{Attn}}^l_{1:T,1:h,:,1:HW} \right].
    \label{eq16}
\end{align}
Within each AdaS-Block, the S-LGU or S-FFN module conducts adaptive inference based on AdaE-IF, while the importance of spiking token is measured precisely by the attention scores of the preceding S-GSA. As shown in Eq.\eqref{eq16}, we uniformly average $\mathbf{Attn}^l$ along the temporal, attention-head and token-column dimensions to obtain the overall scoring of spiking tokens $\tilde{\mathbf{Attn}}^l_{1:T,1:h,:,1:HW}\in\mathbb{R}^{B\times HW}$, where the tokens with top-ranked scores are regarded as salient representation regions and routed to the fine-grained firing pattern.
\section{Experiments}
\begin{table*}[t]
    \caption{\textbf{Comparison of representative SNN approaches across learning paradigms and network architectures on ImageNet-1k.} Here \textbf{T} denotes the inference time-steps, \textbf{Energy} is the estimated inference energy per image, computed following the convention of \cite{zhou2023spikformer}. \textbf{\textcolor{ratioaeDeep}{Ratio AE}} quantifies the accuracy delivered per unit of energy cost, where a higher value indicates better energy efficiency.}
    \renewcommand\arraystretch{1.0}
	\centering
    \resizebox{\linewidth}{!}{
      \begin{threeparttable}
	   \begin{tabular}{cc|ccccc>{\columncolor{ratioae}}c} \Xhline{1pt}
        \textbf{Method} & \textbf{Type} & \textbf{Architecture} & \textbf{Param.(M)} & \textbf{T} & \textbf{Energy(mJ)} & \textbf{Acc.(\%)} & \textbf{Ratio AE} \\ \hline

        \multirow{2}{*}{Spiking ResNet \cite{hu2021residual}} & \multirow{2}{*}{ANN-SNN Conversion} & ResNet-34 & 21.79 & 350 & 59.30 & 71.61 & 1.21 \\
        & & ResNet-50 & 25.56 & 350 & 70.93 & 72.75 & 1.03 \\ \hline
        
        STBP-tdBN \cite{zheng2021going} & STBP Training & Spiking ResNet-34 & 21.79 & 6 & 6.39 & 63.72 & 9.97 \\
        \multirow{4}{*}{SEW ResNet \cite{fang2021deep}} & \multirow{4}{*}{STBP Training} & SEW ResNet-34 & 21.79 & 4 & 4.04 & 67.04 & 16.59 \\
        & & SEW ResNet-50 & 25.56 & 4 & 4.89 & 67.78 & 13.86 \\
        & & SEW ResNet-101 & 44.55 & 4 & 8.91 & 68.76 & 7.72 \\
        & & SEW ResNet-152 & 60.19 & 4 & 12.89 & 69.26 & 5.37 \\
        \multirow{2}{*}{MS ResNet \cite{hu2024residual}} & \multirow{2}{*}{STBP Training} & MS ResNet-34 & 21.79 & 6 & 4.29 & 69.42 & 16.18 \\
        & & MS ResNet-104\tnote{$\dag$} & 77.28 & 5 & 10.19 & 76.02 & 7.46 \\
        GAC-SNN \cite{qiu2024gated} & STBP Training & MS ResNet-34 & 21.93 & 6 & 3.38 & 70.42 & 20.83 \\ 
        Att-MS ResNet \cite{yao2023attention} & STBP Training & MS ResNet-104\tnote{$\dag$} & 78.37 & 4 & 7.30 & 77.08 & 10.56 \\ \hline

        \multirow{3}{*}{Spikformer \cite{zhou2023spikformer}} & \multirow{3}{*}{STBP Training} & Spikformer-8-384 & 16.81 & 4 & 7.73 & 70.24 & 9.09 \\
        & & Spikformer-8-512 & 29.68 & 4 & 11.58 & 73.38 & 6.34 \\
        & & Spikformer-8-768 & 66.34 & 4 & 21.48 & 74.81 & 3.48 \\
        \multirow{3}{*}{SDT \cite{yao2023spike}} & \multirow{3}{*}{STBP Training} & SDT-8-384 & 16.81 & 4 & 3.90 & 72.28 & 18.53 \\
        & & SDT-8-512 & 29.68 & 4 & 4.50 & 74.57 & 16.57 \\
        & & SDT-8-768 & 66.34 & 4 & 6.09 & 76.32 & 12.53 \\ 
        \multirow{2}{*}{Meta-Sformer \cite{yao2024spike}} & \multirow{2}{*}{STBP Training} & MetaSformer-256/320 & 15.1 & 4 & 16.7 & 74.1 & 4.44 \\
        & & MetaSformer-384/480 & 31.3 & 4 & 32.8 & 77.2 & 2.35 \\  
        \multirow{2}{*}{SNN-ViT \cite{wang2025svit}} & \multirow{2}{*}{STBP Training} & SViT-8-256 & 13.7 & 4 & 14.28 & 74.66 & 5.23 \\
        & & SViT-8-384 & 30.4 & 4 & 20.83 & 76.87 & 3.69 \\        
        \multirow{3}{*}{Spikingformer \cite{zhou2026spikingformer}} & \multirow{3}{*}{STBP Training} & Spikingformer-8-384 & 16.81 & 4 & 5.61 & 74.35 & 13.25 \\
        & & Spikingformer-8-512 & 29.68 & 4 & 8.68 & 76.54 & 8.82 \\
        & & Spikingformer-8-768 & 66.34 & 4 & 16.30 & 77.64 & 4.76 \\ \hline

        \multirow{3}{*}{\textbf{This Work}} & \multirow{3}{*}{\textbf{\makecell{AdaS-ViT Training}}} & AdaS-ViT Small & \textbf{5.36} & 4 & \textbf{1.02} & \textbf{74.03} & \textbf{72.58} \\
        & & AdaS-ViT Base & \textbf{9.40} & 4 & \textbf{1.66} & \textbf{77.25} & \textbf{46.54} \\
        & & AdaS-ViT Large & \textbf{14.56} & 4 & \textbf{2.45} & \textbf{78.94} & \textbf{32.22} \\
        
        \Xhline{1pt}
	    \end{tabular}
        \begin{tablenotes}
            \small
            \item[$\dag$] enlarges 288$\times$288 input at inference, remaining cases use 224$\times$224 for both training and inference.
        \end{tablenotes}
        \end{threeparttable}
    }
	\label{tab01}
\end{table*}
In this section, we conduct quantitative performance comparisons of AdaS-ViT against prior SNN works built upon different learning paradigms and network architectures, over a series of benchmarks with diverse data modalities and scales, focusing on core metrics such as inference accuracy and energy consumption. The selected datasets include ImageNet-1k \cite{Deng2009ImageNet}, CIFAR-10(100) \cite{Krizhevsky2009CIFAR100} and CIFAR10-DVS \cite{li2017cifar10}. Tabs.\ref{tab01}-\ref{tab02} correspond to direct training and fine-tuning under unified data-augmentation strategies \cite{DeVries2017,zhang2017mixup,yun2019cutmix,cubuk2020randaugment}, with 300 and 100 training epochs, respectively.
\begin{table*}[t]
    \caption{\textbf{Generalization across data scales and modalities:} comparison with prior SNN learning methods on static (CIFAR-10, CIFAR-100) and neuromorphic (CIFAR10-DVS) benchmarks.
    }
    \renewcommand\arraystretch{1.0}
	\centering
    \resizebox{0.95\linewidth}{!}{
    	\begin{tabular}{cc|cccccc} \Xhline{1pt}
            \multirow{2}{*}{\textbf{Method}} & \multirow{2}{*}{\textbf{Type}} & \multicolumn{2}{c}{\textbf{CIFAR-10}} & \multicolumn{2}{c}{\textbf{CIFAR-100}} & \multicolumn{2}{c}{\textbf{CIFAR10-DVS}} \\ \cline{3-8}
            & & \textbf{Time-step} & \textbf{Acc.(\%)} & \textbf{Time-step} & \textbf{Acc.(\%)} & \textbf{Time-step} & \textbf{Acc.(\%)} \\ \hline
            STBP-tdBN \cite{zheng2021going} & STBP Training & 4 & 92.92 & - & - & 10 & 67.8 \\
            TET \cite{deng2022temporal} & STBP Training & 4 & 94.44 & 4 & 74.47 & 10 & 83.17 \\
            QCFS \cite{bu2022optimal} & ANN-SNN Conversion & 8 & 94.82 & 8 & 78.48 & - & - \\
            SLTT \cite{Meng2023SLTT} & Online Training & 6 & 94.44 & 6 & 74.38 & 10 & 82.20 \\
            GAC-SNN \cite{qiu2024gated} & STBP Training & 4 & 96.24 & 4 & 79.83 & - & - \\ \hline

            \multirow{2}{*}{Spikformer \cite{zhou2023spikformer}} & STBP Training & 4 & 95.19 & 4 & 77.86 & 16 & 80.6 \\
            & Transfer Learning & 4 & 97.03 & 4 & 83.83 & - & - \\
            SDT \cite{yao2023spike} & STBP Training & 4 & 95.6 & 4 & 78.4 & 16 & 80.0 \\
            SNN-ViT \cite{wang2025svit} & STBP Training & - & 96.1 & - & 80.1 & - & 82.3 \\
            Spikingformer \cite{zhou2026spikingformer} & STBP Training & 4 & 95.95 & 4 & 80.37 & 16 & 81.4 \\ \hline

            \textbf{This Work} & \textbf{AdaS-ViT Training} & 4 & \textbf{98.56} & 4 & \textbf{89.24} & 4 & \textbf{88.8} \\
            \Xhline{1pt}
    	\end{tabular}
    }
	\label{tab02}
\end{table*}
\subsection{Energy Efficiency Evaluation with Prior Comparative Works}
As shown in Tab.\ref{tab01}, we systematically collect representative methods spanning ANN-SNN Conversion \& STBP training paradigms and covering both S-CNN \& S-ViT architectures as comparison targets. From the perspective of inference efficiency, AdaS-ViT (5.36M) requires only 24.60\%-20.97\% parameters of Spiking ResNet \cite{hu2021residual}, yet substantially outperforms it with a 71$\times$-72$\times$ Ratio AE and an 87.5$\times$ acceleration in inference time-steps, demonstrating a qualitative improvement over vanilla conversion scheme. For the S-CNN family of methods, AdaS-ViT (9.40M) surpasses SEW ResNet-152 \cite{fang2021deep} with 50.79M fewer parameters, 11.23mJ lower energy consumption and 7.99\% higher accuracy. Meanwhile, it further exceeds MS ResNet-104 \cite{hu2024residual} and Att-MS ResNet-104 \cite{yao2023attention} by 39$\times$ and 36$\times$ in Ratio AE while maintaining the accuracy lead.

On the S-ViT architecture, AdaS-ViT (14.56M) keeps its parameter count no larger than those of the counterpart cases, saving 5.28mJ and 3.16mJ of computational power compared with Spikformer-8-384 \cite{zhou2023spikformer} and Spikingformer-8-384 \cite{zhou2026spikingformer} while improving the inference accuracy by 8.70\% and 4.59\%. Against SDT-8-384 \cite{yao2023spike} with linear-complexity attention computation and MetaSformer-256/320 \cite{yao2024spike} based on a convolution-attention hybrid architecture, our scheme achieves accuracy improvements of 6.66\% and 4.84\% together with Ratio AE gains of 14$\times$-28$\times$.
\begin{figure*}[t]
  \centering
  \includegraphics[width=\textwidth]{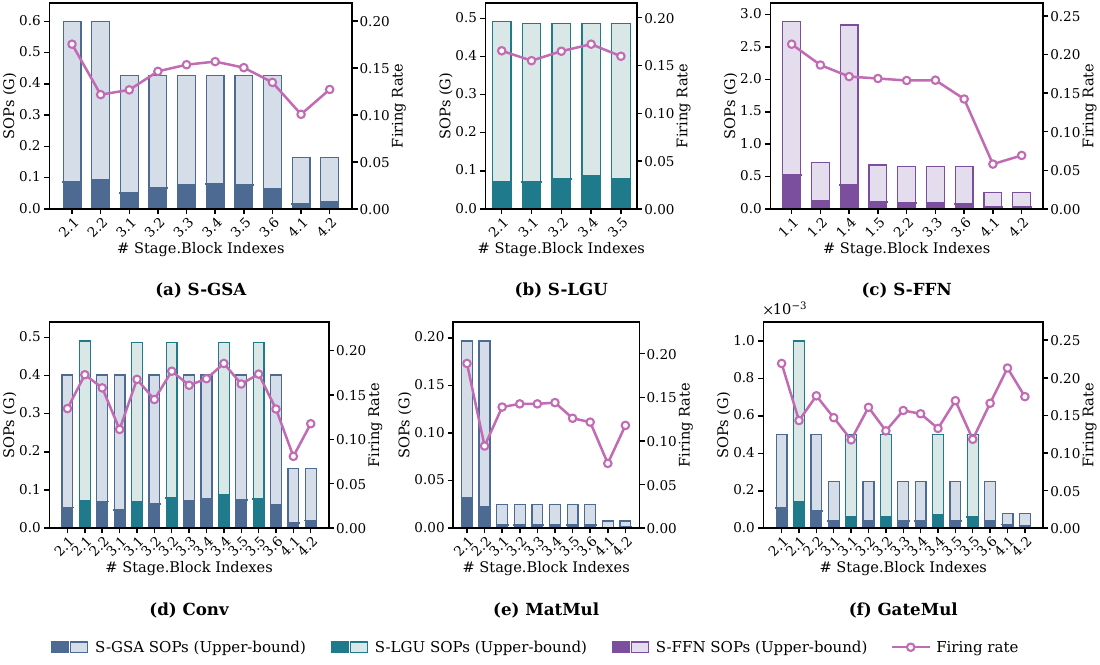}
  \caption{\textbf{Fine-grained energy breakdown of AdaS-ViT Large on ImageNet-1k.}
  Module-level (a)--(c) and Operator-level (d)--(f) statistics of SOPs and average spike firing rates.
  For each block, Bar hues identify the module family, the total height of a bar denotes the theoretical SOPs Upper-bound while its dark lower segment denotes the actually incurred SOPs, the curve on the right axis reports the average firing rate.
  }
  \label{fig04}
\end{figure*}
\subsection{Generalization Learning of AdaS-ViT}
The generalization ability of pre-trained AdaS-ViT backbone is further verified on downstream benchmarks with static and neuromorphic data. We summarize various prior SNN works that adopt conversion-based, STBP-based, online training and transfer learning strategies in Tab.\ref{tab02}. It can be observed that our method establishes a consistent performance advantage in all cases. For CIFAR-10, we lead the best experimental cases based on S-CNN and S-ViT in the table by 2.32\% and 1.53\%, while on CIFAR-100 the corresponding advantages further expand to 9.41\% and 5.41\%. For CIFAR10-DVS, the accuracy achieved by our adaptive spike firing mechanism within only 4 time-steps is 5.63\% higher than the corresponding performance of TET \cite{deng2022temporal} with 10 time-steps.

\subsection{Distribution Details for Energy Consumption}
\textbf{Module-level.} Fig.~\ref{fig04}(a)-(c) show that the incurred SOPs of all three module families remain far below their dense upper-bounds across stages. S-GSA and S-LGU incur per-block SOPs of 0.0145-0.0925 G and 0.0699-0.0867 G against upper-bounds of 0.1646-0.5986 G and about 0.49 G, so adaptive firing activates only 8.8\%-18.7\% of potential synaptic operations, saving 81\%-91\% of the dense energy budget. The corresponding firing rates are confined to 0.10-0.18 for S-GSA and 0.15-0.17 for S-LGU, then drop to 0.058-0.069 for the Stage-IV S-FFN blocks, confirming that progressive learning and routing jointly suppress spike activity in deeper, lower-resolution stages. The stem S-FFN blocks (1.1-1.5) incur the largest absolute SOPs of 0.5262 G and 0.3761 G due to full-resolution inputs, yet still occupy only 18.2\% and 13.2\% of their upper-bounds.

\textbf{Operator-level.} Fig.~\ref{fig04}(d)-(f) decompose the SOPs of S-GSA and S-LGU into convolution, attention matrix multiplication (MatMul) and gating element-wise product (GateMul). Convolution dominates the operator budget in both modules, \eg, 0.0483 G versus 0.0034 G of MatMul in block 3.1 of S-GSA, while the MatMul cost is curtailed by the S-GSA mechanism, falling by roughly one order of magnitude from Stage II (0.022-0.032 G) through Stage III (about 0.0035 G) to below 0.001 G in Stage IV. Most importantly, the SOPs of GateMul within each module are merely at the level of $10^{-5}$-$10^{-4}$ G, \ie, less than 0.2\% of its host block, verifying that routing is essentially free in energy, and the moderate gating-neuron firing rates of 0.12-0.22 indicate that the gate selectively passes salient tokens rather than firing densely.
\section{Conclusions}
We present AdaS-ViT, a test-time adaptive-routing S-ViT built upon the AdaE-IF model, which selectively routes spiking tokens to coarse- or fine-grained spike firing patterns, together with the S-GSA \& S-LGU modules and a progressive spatio-temporal curriculum. Extensive experiments across static and neuromorphic benchmarks demonstrate its consistent advanced performance. By reconciling high representation capability with lightweight spike activity, this work paves the way toward ultra-high energy efficiency inference for edge-side computing.

\section*{LLM Usage Statement}
In this work, we use generative AI tools exclusively for editing and refining the manuscript text, code and figures. All AI-assisted sections are thoroughly reviewed by the authors, who take full responsibility for the final content of this work.

{
    \small
    \bibliographystyle{ieeenat_fullname}
    \bibliography{main}
}


\end{document}